\documentclass{article}
\usepackage{ijcai18}

\usepackage{times}
\usepackage{xcolor}
\usepackage{soul}
\usepackage[utf8]{inputenc}
\usepackage[small]{caption}
\usepackage{epsfig}
\usepackage{graphicx}
\usepackage{algorithm}
\usepackage{algorithmic}
\usepackage{amsmath}
\usepackage{amssymb}
\usepackage{bm}
\newcommand{\etal}{\textit{et~al.~}}

\title{Unpaired Multi-Domain Image Generation via Regularized Conditional GANs}

\author{
Xudong Mao \and
Qing Li
\\ 
Department of Computer Science, City University of Hong Kong \\
xudong.xdmao@gmail.com,
itqli@cityu.edu.hk
}

\begin{document}

\maketitle

\begin{abstract}
In this paper, we study the problem of multi-domain image generation, the goal of which is to generate pairs of corresponding images from different domains. With the recent development in generative models, image generation has achieved great progress and has been applied to various computer vision tasks. However, multi-domain image generation may not achieve the desired performance due to the difficulty of learning the correspondence of different domain images, especially when the information of paired samples is not given. To tackle this problem, we propose Regularized Conditional GAN (\mbox{RegCGAN}) which is capable of learning to generate corresponding images in the absence of paired training data. RegCGAN is based on the conditional GAN, and we introduce two regularizers to guide the model to learn the corresponding semantics of different domains. We evaluate the proposed model on several tasks for which paired training data is not given, including the generation of edges and photos, the generation of faces with different attributes, etc. The experimental results show that our model can successfully generate corresponding images for all these tasks, while outperforms the baseline methods. We also introduce an approach of applying \mbox{RegCGAN} to unsupervised domain adaptation.
\end{abstract}
\section{Introduction}

\begin{figure*}[t]
\centering
\small
\begin{tabular}{c}
 \includegraphics[width=0.86\textwidth]{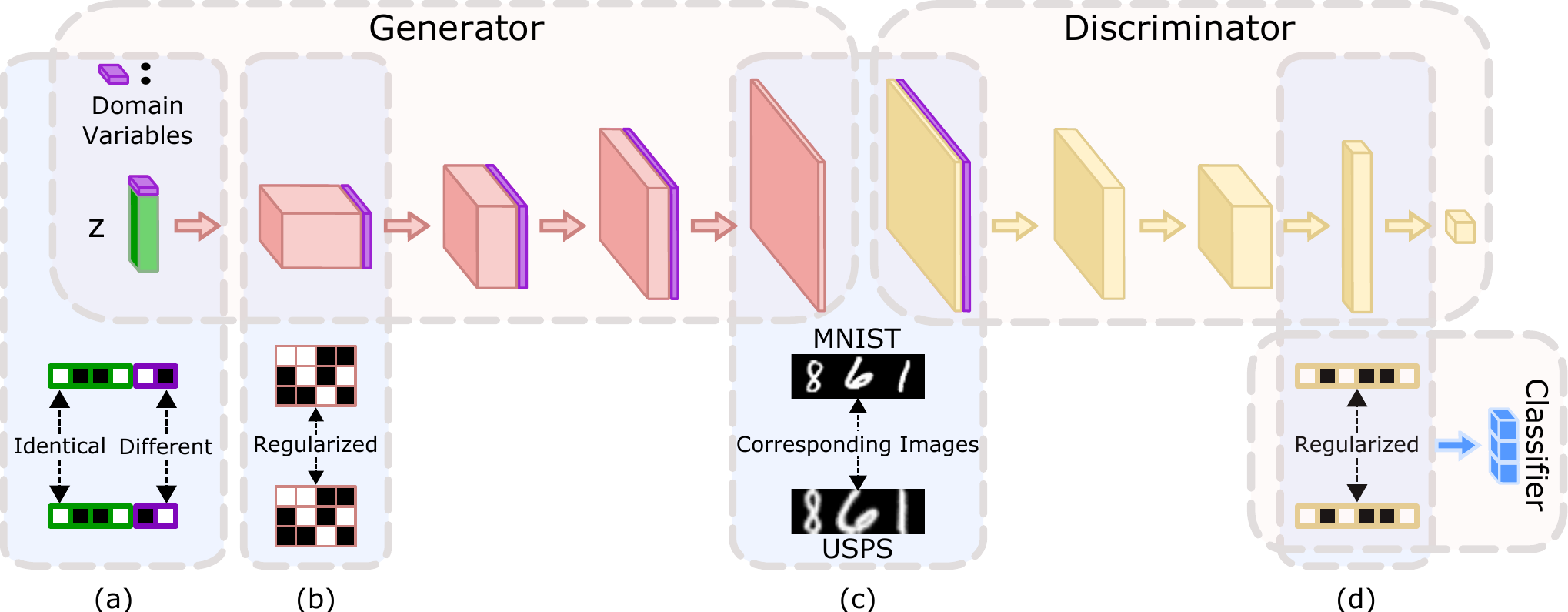}
\end{tabular}
\caption{
The framework of RegCGAN. The domain variables (in purple) are conditioned to all the layers of the generator and the input image layer of the discriminator. (a): The pairs of input consist of identical latent variables but different domain variables. (b): One regularizer is used in the first layer of the generator. It penalizes the distances between the first layer's output of the input pairs in (a), which guides the generator to decode similar high-level semantics for corresponding images. (c): The generator generates pairs of corresponding images. (d): Another regularizer is used in the last hidden layer of the discriminator. This regularizer enforces the discriminator to output similar losses for the corresponding images. These similar losses are used to update the generator. This regularizer also makes the model output invariant feature representations for the corresponding images from different domains. The learned invariant feature representations can be used for domain adaptation by attaching a classifier.
}
\label{fig:arch}
\end{figure*}
Multi-domain image generation is an important extension of image generation in computer vision. It has many promising applications such as improving the generated image quality ~\cite{Dosovitskiy2015,Wang2016}, image-to-image translation ~\cite{Perarnau2016,Wang2017}, and unsupervised domain adaptation ~\cite{Liu2016}. As shown in Figures \ref{fig:digits} and \ref{fig:edge}, a successful model for multi-domain image generation should be able to generate pairs of corresponding images which share common semantics but are of different domain-specific semantics. Several early approaches ~\cite{Dosovitskiy2015,Wang2016} have been proposed, but they are all in the supervised setting, which means that they require the information of paired samples to be available. In practice, however, building paired training datasets can be very expensive and may not always be feasible.

Recently, CoGAN ~\cite{Liu2016} has been proposed and achieved great success in multi-domain image generation. In particular, CoGAN models the problem as to learn a joint distribution over multi-domain images by coupling multiple GANs. Unlike previous methods that require paired training data, CoGAN is able to learn the joint distribution without any paired samples. However, it falls short for some difficult tasks such as the generation of edges and photos, as demonstrated by experiments. 

In this paper, we propose a new framework called Regularized Conditional GAN (RegCGAN). Like CoGAN, RegCGAN is also capable of performing multi-domain image generation in the absence of paired samples. RegCGAN is based on the conditional GAN ~\cite{Mirza2014} and tries to learn a conditional distribution over multi-domain images, where the domain-specific semantics are encoded in the conditioned domain variables, and the common semantics are encoded in the shared latent variables.

As pointed out in ~\cite{Liu2016}, directly using conditional GAN will fail to learn the corresponding semantics. To overcome this problem, we propose two regularizers to guide the model to encode the common semantics in the shared latent variables, which in turn makes the model to generate corresponding images. As shown in Figure \ref{fig:arch}(a)(b), one regularizer is used in the first layer of the generator. This regularizer penalizes the distances between the first layer's output of the paired input, where the paired input should consist of identical latent variables but different domain variables. As a result, it enforces the generator to decode similar high-level semantics for the paired input, since the first layer decodes the highest level semantics. This strategy is based on the fact that corresponding images from different domains always share some high-level semantics (ref. Figures \ref{fig:digits}, \ref{fig:edge}, and \ref{fig:face}). As shown in Figure \ref{fig:arch} (c)(d), the second regularizer is added to the last hidden layer of the discriminator which is responsible for encoding the highest level semantics. This regularizer enforces the discriminator to output similar losses for the pairs of corresponding images. These similar losses are then used to update the generator, which guides the generator to generate similar (corresponding) images.

One intuitive application of RegCGAN is unsupervised domain adaptation, since the second regularizer (Figure \ref{fig:arch}(d)) is able to make the last hidden layer to output invariant feature representations for corresponding images (Figure \ref{fig:arch}(c)). We can attach a classifier to the last hidden layer, and the classifier is jointly trained with the discriminator using the labeled images from the source domain. As a result, the classifier is able to classify the images from the target domain due to the learned invariant feature representations.

\section{Related Work}
\subsection{Multi-Domain Image Generation}
Image generation is one of the most fundamental problems in computer vision. Classic approaches include Restricted Boltzmann Machine ~\cite{Tieleman2008} and Autoencoder~\cite{Bengio2013}. Recently, two successful approaches, Variational Autoencoder (VAE)~\cite{Kingma2013} and Generative Adversarial Network (GAN) ~\cite{Goodfellow2014}, have been proposed. Our model in this paper is based on GAN. The idea of GAN is to find the Nash Equilibrium between the generator network and discriminator network. GAN has achieved great success in image generation, and numerous variants ~\cite{Radford2015,Nowozin2016,Arjovsky2017,Mao2017} have been proposed for improving the image quality and training stability. 

Multi-domain image generation is an extension problem of image generation in which two or more domain images are provided. A successful model should be able to generate pairs of corresponding images, which means that the image pairs share some common semantics but are of different domain-specific semantics. It has many promising applications such as improving the generated image quality ~\cite{Dosovitskiy2015,Wang2016} and image-to-image translation ~\cite{Perarnau2016,Wang2017}.  Early approaches ~\cite{Dosovitskiy2015,Wang2016} are under the supervised setting, where the information of paired images is provided. However, building training datasets with paired information is not always feasible and can be very expensive. The recent proposed CoGAN ~\cite{Liu2016} is able to perform multi-domain image generation in the absence of any paired images. CoGAN consists of multiple GANs and each GAN corresponds to one image domain. Furthermore, the weights of some layers are tied to learn the shared semantics.

\subsection{Regularization Methods}
Regularization methods have been proven to be effective in GAN learning ~\cite{Che2016,Gulrajani2017,Roth2017}. Che \etal~\shortcite{Che2016} introduced several types of regularizers which penalize the missing modes. These regularizers are able to relieve the missing modes problem. Gulrajani \etal ~\shortcite{Gulrajani2017} proposed an effective way of regularizing the gradients of the points sampled between the data distribution and the generator distribution. Moreover, Roth \etal ~\shortcite{Roth2017} proposed a weighted gradient-based regularizer which can be applied to various GANs. In this paper, we adopt the regularization method for enforcing the model to generate corresponding images.

\begin{figure*}[t]
\centering
\small
\begin{tabular}{ccc}
 \includegraphics[width=0.31\textwidth]{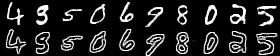}
&
 \includegraphics[width=0.31\textwidth]{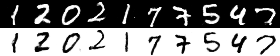}
& 
 \includegraphics[width=0.31\textwidth]{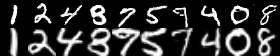}

\\
(a) Digits and edge digits.
&
(b) Digits and negative digits.
&
(c) MNIST and USPS digits.
\end{tabular}
\caption{
Generated image pairs on digits.
}
\label{fig:digits}
\end{figure*}

\section{Framework and Approach}
\subsection{Generative Adversarial Network}
The framework of GAN consists of two roles, the discriminator $D$ and the generator $G$. Given a data distribution $p_\text{data}$, $G$ tries to learn the distribution $p_g$ over data $\bm{x}$. $G$ starts from sampling the noise input $\bm{z}$ from a simple distribution $p_{z}(\bm{z})$, and then maps $\bm{z}$ to data space $G(\bm{z}; \theta_g)$. On the other hand, $D$ aims to distinguish whether a sample is from $p_\text{data}$ or from $p_g$.  The objective for GAN can be formulated as follows:

\begin{equation}
\label{eq:gan}
\begin{split}
\min_G \max_D V(G, D) =& \mathbb{E}_{\bm{x} \sim p_{\text{data}}(\bm{x})}[\log D(\bm{x})] + \\
& \mathbb{E}_{\bm{z} \sim p_{\bm{z}}(\bm{z})}[\log (1 - D(G(\bm{z})))].
\end{split}
\end{equation}

\subsection{Regularized Conditional GAN}
\label{sec:regcgan}
In our approach, the problem of multi-domain image generation is modeled as to learn a conditional distribution $p_{\text{data}}(\bm{x}|d)$ over data $\bm{x}$, where $d$ denotes the domain variable. We propose the Regularized Conditional GAN (RegCGAN) for learning $p_{\text{data}}(\bm{x}|d)$. Our idea is to encode the domain-specific semantics in the domain variable $d$ and to encode the common semantics in the shared latent variables $\bm{z}$. To achieve this, the conditional GAN is adopted and two regularizers are proposed. One regularizer is added to the first layer of the generator, and the other one is added to the last hidden layer of the discriminator. 

Specifically, as Figure \ref{fig:arch} shows, for an input pair $(\bm{z}d_i, \bm{z}d_j)$ with identical $\bm{z}$ but different $d$, the first regularizer penalizes the distance between the first layer's output of $\bm{z}d_i$ and $\bm{z}d_j$, which enforces $G$ to decode similar high-level semantics,  since the first layer decodes the highest level semantics. On the other hand, for a pair of corresponding images $(\bm{x}_i, \bm{x}_j)$, the second regularizer penalizes the distance between the last layer's output of $\bm{x}_i$ and $\bm{x}_j$. As a result, $D$ outputs similar losses for the pairs of corresponding images. When updating $G$, these similar losses guide $G$ to generate similar (corresponding) images. Note that to use the above two regularizers, it requires constructing pairs of input which are of identical $\bm{z}$ but different $d$.

Formally, when training, we construct mini-batches with pairs of input $(\bm{z},d=0)$ and $(\bm{z},d=1)$, where the noise input $\bm{z}$ is the same. $G$ maps the noise input $\bm{z}$ to a conditional data space $G(\bm{z} | d)$. An L2-norm regularizer is used to enforce $G_{h_0}(\bm{z}|d)$, the output of $G$'s first layer, to be similar for each paired input. Another L2-norm regularizer is used to enforce $D_{h_i}(G(\bm{z} | d))$, the output of $D$'s last hidden layer, to be similar for each paired input. Then the objective function for RegCGAN can be formulated as follows: 

\begin{alignat}{2}
\label{eq:rcgan_G}
&\min_G \max_D V&&(G, D) = \mathcal{L}_{\text{GAN}}(G,D)+\lambda \mathcal{L}_{\text{reg}}(G)+\beta \mathcal{L}_{\text{reg}}(D), \nonumber \\
&\mathcal{L}_{\text{GAN}}(G,D&&) = \mathbb{E}_{\bm{x} \sim p_{\text{data}}(\bm{x}|d)}[\log D(\bm{x}|d)] + \nonumber \\
& && \quad \quad  \mathbb{E}_{\bm{z} \sim p_{\bm{z}}(\bm{z})}[\log (1 - D(G(\bm{z}|d)))], \nonumber \\
& \mathcal{L}_{\text{reg}}(G)=&&  \mathbb{E}_{\bm{z} \sim p_{\bm{z}}(\bm{z})} [\|G_{h_0}(\bm{z} | d=0) - G_{h_0}(\bm{z} | d=1)\|^2],\nonumber \\
 &\mathcal{L}_{\text{reg}}(D)=&&  \mathbb{E}_{\bm{z} \sim p_{\bm{z}}(\bm{z})} [-\|D_{h_i}(G(\bm{z} | d=0)) -  \nonumber \\
 &  && \qquad \qquad \qquad \quad D_{h_i}(G(\bm{z}|  d=1))\|^2],
\end{alignat}
where the scalars $\lambda$ and $\beta$ are used to adjust the weights of the regularization terms, $\|\cdot\|$ denotes the $l^2$-norm, $d=0$ and $d=1$ denote the source domain and target domain, respectively, $G_{h_0}(\cdot)$ denotes the output of $G$'s first layer, and $D_{h_i}(\cdot)$ denotes the output of $D$'s last hidden layer.

As stated before, RegCGAN can be applied to unsupervised domain adaptation, since the last hidden layer of $D$ is able to output invariant feature representations for the corresponding images from different domains. Based on the invariant feature representations, we attach a classifier to the last hidden layer of $D$. The classifier is jointly trained with $D$, and the joint objective function is:

\begin{equation}
\label{eq:uda}
\begin{split}
\min_{G,C} \max_D V(G, D, C) = \mathcal{L}_{\text{GAN}}(G,D)+\lambda \mathcal{L}_{\text{reg}}(G)  \\
+\beta \mathcal{L}_{\text{reg}}(D)+ \gamma \mathcal{L}_{\text{cls}}(C), \\
\end{split}
\end{equation}
where the scalars $\lambda$, $\beta$, and $\gamma$ are used to adjust the weights of the regularization terms and the classifier, and $\mathcal{L}_{\text{cls}}(C)$ is a typical cross-entropy loss.

Note that our approach to domain adaptation is different from the method used in ~\cite{Ganin2016} which tries to minimize the difference between the overall distribution of the source and target domains. In contrast, the minimization of our approach is among samples belonging to the same category, because we only penalize the distances between the pairs of corresponding images which belong to the same category.

\begin{figure*}[t]
\centering
\small
\begin{tabular}{cc}
 \includegraphics[width=0.46\textwidth]{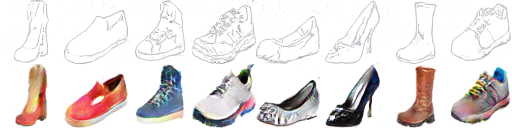} & \includegraphics[width=0.46\textwidth]{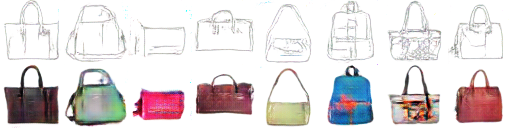}
\\
(a) Shoes by RegCGAN (Ours). & (b) Handbags by RegCGAN (Ours).
\\
 \includegraphics[width=0.46\textwidth]{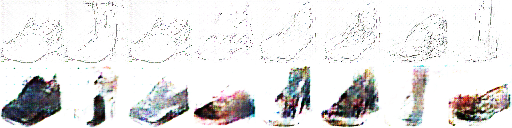} & \includegraphics[width=0.46\textwidth]{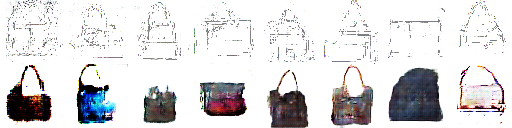}
\\
(c) Shoes by CoGAN. & (d) Handbags by CoGAN.
\end{tabular}
\caption{
Generated image pairs on shoes and handbags.
}
\label{fig:edge}
\end{figure*}

\section{Experiments}
\subsection{Implementation Details}
Except for the tasks about digits (i.e., MNIST and USPS), we adopt LSGAN ~\cite{Mao2017} for training the models due to the fact that LSGAN generates higher quality images and perform more stably. For digits tasks we still adopt standard GAN since we find that LSGAN will sometimes generate unaligned digit pairs.

We use Adam optimizer with the learning rates of $0.0005$ for LSGAN and $0.0002$ for standard GAN. For the hyper-parameters in Equations \ref{eq:rcgan_G} and \ref{eq:uda}, we set $\lambda=0.1$, $\beta=0.004$, and $\gamma=1.0$ found by grid search. Our implementation is available at https://github.com/xudonmao/RegCGAN.

\subsection{Digits}
\label{sec:digits}
We first evaluate RegCGAN on MNIST and USPS datasets. Since the image sizes of MNIST and USPS are different, we resize the images in USPS to the same resolution (i.e., $28 \times 28$) of MNIST. We train RegCGAN for the following three tasks. Following literature ~\cite{Liu2016}, the first two tasks are to perform the generations of 1) digits and edge digits; 2) digits and negative digits. The third one is to perform the generation of MNIST and USPS digits. For these tasks, we design the network architecture following the suggestions in ~\cite{Radford2015}, where the generator consists of four transposed convolutional layers and the discriminator is a variant of LeNet ~\cite{Lecun98}. The generated image pairs are shown in Figure \ref{fig:digits}, where we can see clearly that RegCGAN succeeds to generate corresponding digits for all the three tasks.

\begin{figure}[t]
\centering
\small
\begin{tabular}{cc}
 \includegraphics[width=0.22\textwidth]{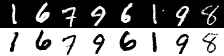}
&
 \includegraphics[width=0.22\textwidth]{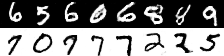}
\\
(a) With regularizer.
&
(b) Without regularizer.
\end{tabular}
\caption{
Comparison experiments between the models with and without the regularizer.
}
\label{fig:wo_reg}
\end{figure}

\vspace{2pt}
\noindent\textbf{Without Regularizer} If we remove the proposed regularizers in RegCGAN, the model will fail to generate corresponding digits as Figure \ref{fig:wo_reg} shows. This demonstrates that the proposed regularizers play an important role in generating corresponding images. 

\subsection{Edges and Photos}
\label{sec:edge}
We also train RegCGAN for the task of generating corresponding edges and photos. The Handbag ~\cite{Zhu2016} and Shoe ~\cite{Yu2014} datasets are used for this task. We randomly shuffle the edge images and realistic photos to avoid utilizing the pair information. We resize all the images to a resolution of $64 \times 64$. For the network architecture, both the generator and the discriminator consist of four transposed/strided-convolutional convolutional layers. As shown in Figure \ref{fig:edge}(a)(b), RegCGAN is able to generate corresponding images of edges and photos.  

\vspace{2pt}
\noindent\textbf{Comparison with CoGAN} We also train CoGAN, which is the current state-of-the-art method, on edges and photos using the official implementation of CoGAN. We evaluate two network architectures for CoGAN: (1) the architecture used in CoGAN ~\cite{Liu2016} and (2) the same architecture to RegCGAN. We also evaluate the standard GAN loss and least squares loss (LSGAN) for CoGAN. But all of these settings fail to generate corresponding images of edges and photos. The results are shown in Figure \ref{fig:edge}(c)(d).

\begin{figure*}[t]
\centering
\small
\begin{tabular}{cc}
 \includegraphics[width=0.46\textwidth]{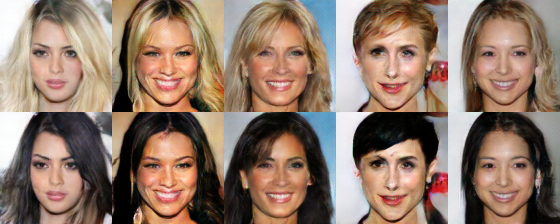} & \includegraphics[width=0.46\textwidth]{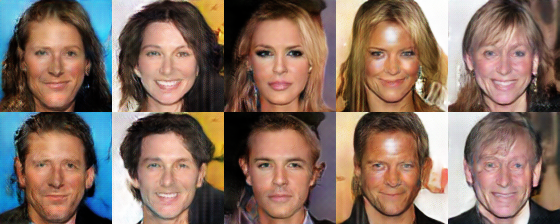}
\\
(a) Blond and black hair by RegCGAN (Ours). & (b) Female and male by RegCGAN (Ours).
\\
 \includegraphics[width=0.46\textwidth]{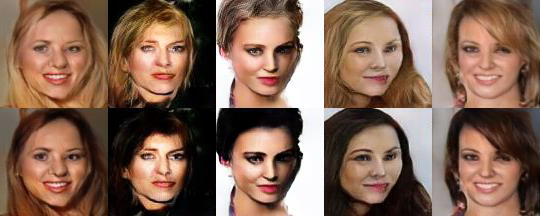} & \includegraphics[width=0.46\textwidth]{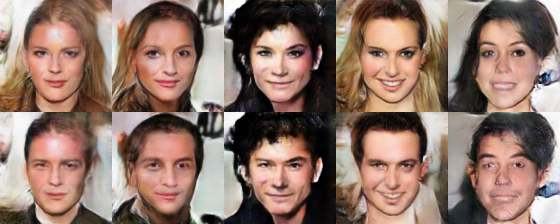} \\
 (c) Blond and black hair by CoGAN. & (d) Female and male by CoGAN.
\end{tabular}
\caption{
Generated image pairs on faces with different attributes. The image pairs of black and blond hair by CoGAN are duplicated from the CoGAN paper.
}
\label{fig:face}
\end{figure*}

\subsection{Faces}
\label{sec:face}
In this task, we evaluate RegCGAN on the CelebA dataset ~\cite{Liu2014}. We first apply a pre-processing method to crop the facial region in the center of the images ~\cite{Karras2017}, and then resize all the cropped images to a resolution of $112 \times 112$. The network architecture used in this task is similar to the one in Section \ref{sec:edge} except for the output dimensions of the layers. We investigate the following two tasks: 1) female with blond and black hair; and 2) female and male. The results are presented in Figure \ref{fig:face}(a)(b). We observe that RegCGAN is able to generate corresponding face images with different attributes, and the corresponding faces are of very similar appearances.

\vspace{2pt}
\noindent\textbf{Comparison with CoGAN} The generated image pairs by CoGAN are also presented in Figure \ref{fig:face}, where the image pairs of black and blond hair by CoGAN are duplicated from ~\cite{Liu2016}. We observe that the image pairs generated by RegCGAN are more consistent and of better quality than the ones by CoGAN, especially for the task of female and male, which is more difficult than the task of blond and black hair.

\begin{figure}[t]
\centering
\small
\begin{tabular}{c}
 \includegraphics[width=0.47\textwidth]{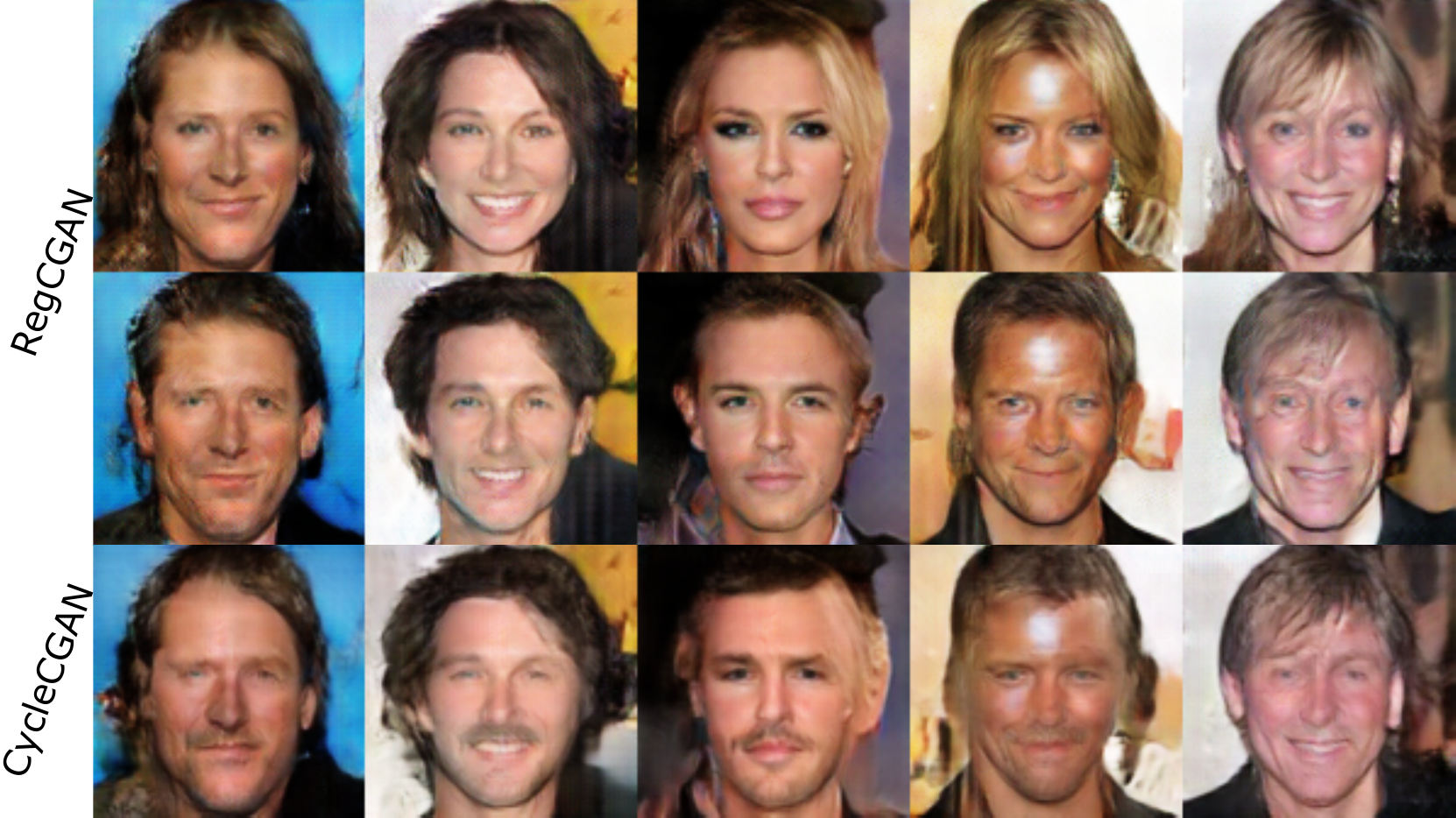}
\end{tabular}
\caption{
Comparison results between RegCGAN and CycleGAN for the task of female and male. The top two rows are generated by RegCGAN. The third row is generated by CycleGAN using the first row as input.
}
\label{fig:cyclegan}
\end{figure}

\vspace{2pt}
\noindent\textbf{Comparison with CycleGAN} We also compare RegCGAN with CycleGAN ~\cite{Zhu2017} which is the state-of-the-art method in image-to-image transition. To compare with CycleGAN, we first generate some image pairs using RegCGAN and then use the generated images in one domain as the input for CycleGAN. The results are presented in Figure \ref{fig:cyclegan}. Compared with RegCGAN, CycleGAN introduces some blur to the generated images. Moreover, the color of the image pairs by RegCGAN is more consistent than the ones by CycleGAN. 

\subsection{Quantitative Evaluation}
To further evaluate the effectiveness of RegCGAN, we conduct a user study on Amazon Mechanical Turk (AMT). For this evaluation, we also use the task of the female and male generation. In particular, given two image pairs randomly selected from RegCGAN and CoGAN, the AMT annotators are asked to choose a better one based on the image quality, perceptual realism, and appearance consistency of female and male. With $3,000$ votes totally, a majority of the annotators preferred the image pairs from RegCGAN in $77.6\%$, demonstrating that the overall image quality of our model is better than the one of CoGAN.

\begin{table}[t]
\renewcommand{\arraystretch}{1.2}
\begin{tabular}{ccc}
\hline
 & CoGAN & RegCGAN (Ours)\\
\hline
User Choice& $673$ / $3000$ ($22.4\%$)&$\textbf{2327}$ / $3000$ ($\textbf{77.6\%}$)\\
\hline
\end{tabular}
\centering
\caption{A user study on the task of female and male generation. With $3,000$ votes totally, $77.6\%$ of the annotators preferred the image pairs from RegCGAN.}
\label{tab:amt}
\end{table}

\subsection{More Applications}
\begin{figure}[t]
\centering
\small
\begin{tabular}{c}
 \includegraphics[width=0.46\textwidth]{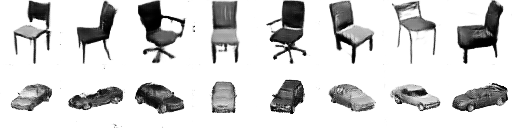}
 \\
(a): Chairs and cars.
 \\
  \includegraphics[width=0.46\textwidth]{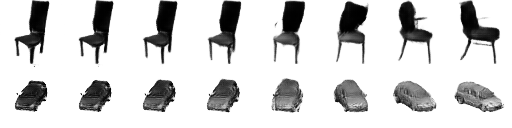}
  \\
(b): Interpolation between two random points in noise input.
\end{tabular}
\caption{
Generated image pairs on chairs and cars, where the orientations are highly correlated. 
}
\label{fig:chair}
\end{figure}

\vspace{2pt}
\noindent\textbf{Chairs and Cars}
In this task, we use two visually completely different datasets, Chairs ~\cite{Aubry2014} and Cars ~\cite{Fidler2012}. Both datasets contain synthesized samples with different orientations. We train RegCGAN on these two datasets to study whether it is able to generate corresponding images sharing the same orientations. The generated results are shown in Figure \ref{fig:chair}, where the image resolution is $64\times 64$. We further perform interpolation between two random points in the latent space as shown in Figure \ref{fig:chair}(b). The interpolation shows smooth transitions of chairs and cars both in viewpoint and style, while the chairs and cars keep facing the same direction.

\begin{figure}[t]
\centering
\small
\begin{tabular}{c}
 \includegraphics[width=0.46\textwidth]{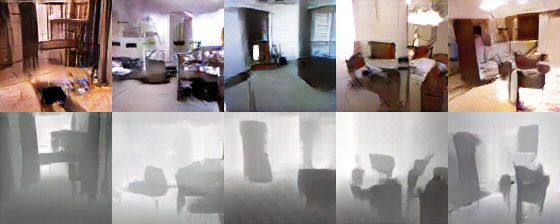}
\end{tabular}
\caption{
Generated image pairs on photos and depth images.
}
\label{fig:depth}
\end{figure}

\vspace{2pt}
\noindent\textbf{Photos and Depths}
\label{sec:depth}
The NYU depth dataset ~\cite{Silberman2012} is used for learning a RegCGAN over photos and depth images. In this task, we first resize all the images to a resolution of $120\times 160$ and then randomly crop $112\times 112$ patches for training. Figure \ref{fig:depth} shows the generated image pairs.

\begin{figure}[t]
\centering
\begin{tabular}{c}
 \includegraphics[width=0.46\textwidth]{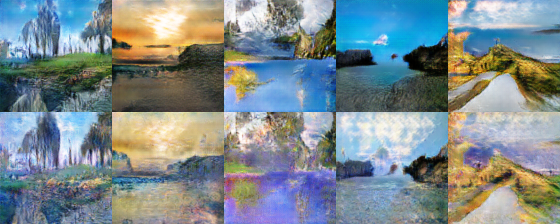}
\end{tabular}
\caption{
Generated image pairs on photos and Monet-style images.
}
\label{fig:monet}
\end{figure}

\vspace{2pt}
\noindent\textbf{Photos and Monet-Style Images}
\label{sec:monet}
In this task we train RegCGAN on the Monet-style dataset ~\cite{Zhu2017}. We use the same pre-processing method as in Section \ref{sec:depth}. Figure \ref{fig:monet} shows the generated image pairs.

\begin{figure}[t]
\centering
\begin{tabular}{c}
 \includegraphics[width=0.46\textwidth]{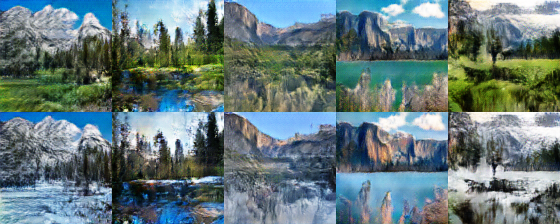}
\end{tabular}
\caption{
Generated image pairs on summer Yosemite and winter Yosemite.
}
\label{fig:summer}
\end{figure}

\vspace{2pt}
\noindent\textbf{Summer and Winter}
We also train RegCGAN on the Summer and Winter dataset ~\cite{Zhu2017}. We use the same pre-processing method as in Section \ref{sec:depth}. Figure \ref{fig:summer} shows the generated image pairs.

\subsection{Unsupervised Domain Adaptation}
\begin{table}[t]
\renewcommand{\arraystretch}{1.2}
\begin{tabular}{ccc}
\hline
Method & MNIST$\to$USPS & USPS$\to$MNIST\\
\hline
&\multicolumn{2}{c}{Evaluated on the sampled set}\\
\hline
DANN&$0.771$&$0.730$\\
ADDA&$0.894\pm0.002$&$\textbf{0.901}\pm0.008$ \\
CoGAN&$0.912\pm0.008$&$0.891\pm0.008$\\
RegCGAN (Ours)&$\textbf{0.931}\pm0.007$&$0.895\pm0.009$\\
\hline
&\multicolumn{2}{c}{Evaluated on the test set}\\
\hline
ADDA&$0.836\pm0.035$&$0.849\pm0.058$\\
CoGAN&$0.882\pm0.018$&$0.822\pm0.081$ \\
RegCGAN (Ours)&$\textbf{0.901}\pm0.009$&$\textbf{0.888}\pm0.015$\\
\hline
\end{tabular}
\centering
\caption{Accuracy results for unsupervised domain adaptation. The top section presents the classification accuracy evaluated on the sampled set of the target domain. The bottom section presents the classification accuracy evaluated on the standard test set of the target domain. The reported accuracies are averaged over $10$ trails with different random samplings.}
\label{tab:uda}
\end{table}

As mentioned in Section \ref{sec:regcgan}, RegCGAN can be applied to unsupervised domain adaptation. In this experiment, MNIST and USPS datasets are used, where one is used as the source domain and the other one is used as the target domain. We set $\lambda=0.1$, $\beta=0.004$, and $\gamma=1.0$ found by grid search. We use the same network architecture as in Section \ref{sec:digits} and attach a classifier at the end of the last hidden layer of the discriminator. Following the experiment protocol in ~\cite{Liu2016,Tzeng2017}, we randomly sample $2,000$ images from MNIST and $1,800$ images from USPS. 

We conduct two comparison experiments between RegCGAN and the baseline methods, including DANN ~\cite{Ganin2016}, ADDA ~\cite{Tzeng2017}, and CoGAN~\cite{Liu2016}. One is to evaluate the classification accuracy directly on the sampled images of the target domain, which is adopted in ~\cite{Liu2016,Tzeng2017}. To further evaluate the generalization error, we further evaluate the classification accuracy on the standard test sets of the target domain. 

The results are presented in Table \ref{tab:uda}. The reported accuracies are averaged over $10$ trails with different random samplings. For the evaluation on the standard test set, RegCGAN significantly outperforms all the baseline methods, especially for the task of USPS to MNIST. This shows that RegCGAN is of smaller generalization error when compared with the baseline methods. For the evaluation on the sampled set, RegCGAN outperforms all the baseline methods for the task of MNIST to USPS, and achieves comparable performance for the task of USPS to MNIST.

\section{Conclusions}
To tackle the problem of multi-domain image generation, we have proposed the Regularized Conditional GAN, where the domain information is encoded in the conditioned domain variables. Two types of regularizers are proposed. One is added to the first layer of the generator, guiding the generator to decode similar high-level semantics. The other one is added to the last hidden layer of the discriminator, enforcing the discriminator to output similar losses for the corresponding images. Various experiments on multi-domain image generation have been conducted. The experimental results show that RegCGAN succeeds to generate pairs of corresponding images for all these tasks, and outperforms all the baseline methods. We have also introduced a method of applying RegCGAN to domain adaptation.

\clearpage

\bibliographystyle{named}
\bibliography{regcgan}

\begin{thebibliography}{}

\bibitem[\protect\citeauthoryear{Arjovsky \bgroup \em et al.\egroup
  }{2017}]{Arjovsky2017}
Martin Arjovsky, Soumith Chintala, and L{\'e}on Bottou.
\newblock Wasserstein gan.
\newblock {\em arXiv:1701.07875}, 2017.

\bibitem[\protect\citeauthoryear{Aubry \bgroup \em et al.\egroup
  }{2014}]{Aubry2014}
Mathieu Aubry, Daniel Maturana, Alexei Efros, Bryan Russell, and Josef Sivic.
\newblock Seeing 3d chairs: exemplar part-based 2d-3d alignment using a large
  dataset of cad models.
\newblock In {\em Computer Vision and Pattern Recognition (CVPR)}, 2014.

\bibitem[\protect\citeauthoryear{Bengio \bgroup \em et al.\egroup
  }{2013}]{Bengio2013}
Yoshua Bengio, Li~Yao, Guillaume Alain, and Pascal Vincent.
\newblock Generalized denoising auto-encoders as generative models.
\newblock {\em arXiv:1305.6663}, 2013.

\bibitem[\protect\citeauthoryear{Che \bgroup \em et al.\egroup
  }{2016}]{Che2016}
Tong Che, Yanran Li, Athul~Paul Jacob, Yoshua Bengio, and Wenjie Li.
\newblock Mode regularized generative adversarial networks.
\newblock {\em arXiv:1612.02136}, 2016.

\bibitem[\protect\citeauthoryear{Dosovitskiy \bgroup \em et al.\egroup
  }{2015}]{Dosovitskiy2015}
Alexey Dosovitskiy, Jost~Tobias Springenberg, and Thomas Brox.
\newblock Learning to generate chairs, tables and cars with convolutional
  networks.
\newblock In {\em Computer Vision and Pattern Recognition (CVPR)}, 2015.

\bibitem[\protect\citeauthoryear{Fidler \bgroup \em et al.\egroup
  }{2012}]{Fidler2012}
Sanja Fidler, Sven Dickinson, and Raquel Urtasun.
\newblock 3d object detection and viewpoint estimation with a deformable 3d
  cuboid model.
\newblock In {\em Advances in Neural Information Processing Systems (NIPS)}.
  2012.

\bibitem[\protect\citeauthoryear{Ganin \bgroup \em et al.\egroup
  }{2016}]{Ganin2016}
Yaroslav Ganin, Evgeniya Ustinova, Hana Ajakan, Pascal Germain, Hugo
  Larochelle, François Laviolette, Mario Marchand, and Victor Lempitsky.
\newblock Domain-adversarial training of neural networks.
\newblock {\em Journal of Machine Learning Research}, 2016.

\bibitem[\protect\citeauthoryear{Goodfellow \bgroup \em et al.\egroup
  }{2014}]{Goodfellow2014}
Ian Goodfellow, Jean Pouget-Abadie, Mehdi Mirza, Bing Xu, David Warde-Farley,
  Sherjil Ozair, Aaron Courville, and Yoshua Bengio.
\newblock Generative adversarial nets.
\newblock In {\em Advances in Neural Information Processing Systems (NIPS)},
  2014.

\bibitem[\protect\citeauthoryear{Gulrajani \bgroup \em et al.\egroup
  }{2017}]{Gulrajani2017}
Ishaan Gulrajani, Faruk Ahmed, Martin Arjovsky, Vincent Dumoulin, and Aaron
  Courville.
\newblock Improved training of wasserstein gans.
\newblock In {\em Advances in Neural Information Processing Systems (NIPS)},
  2017.

\bibitem[\protect\citeauthoryear{Karras \bgroup \em et al.\egroup
  }{2017}]{Karras2017}
Tero Karras, Timo Aila, Samuli Laine, and Jaakko Lehtinen.
\newblock Progressive growing of gans for improved quality, stability, and
  variation.
\newblock {\em arXiv:1710.10196}, 2017.

\bibitem[\protect\citeauthoryear{Kingma and Welling}{2014}]{Kingma2013}
Diederik~P Kingma and Max Welling.
\newblock Auto-encoding variational bayes.
\newblock In {\em International Conference on Learning Representations (ICLR)},
  2014.

\bibitem[\protect\citeauthoryear{Lecun \bgroup \em et al.\egroup
  }{1998}]{Lecun98}
Yann Lecun, Léon Bottou, Yoshua Bengio, and Patrick Haffner.
\newblock Gradient-based learning applied to document recognition.
\newblock In {\em Proceedings of the IEEE}, 1998.

\bibitem[\protect\citeauthoryear{Liu and Tuzel}{2016}]{Liu2016}
Ming-Yu Liu and Oncel Tuzel.
\newblock Coupled generative adversarial networks.
\newblock In {\em Advances in Neural Information Processing Systems (NIPS)},
  2016.

\bibitem[\protect\citeauthoryear{Liu \bgroup \em et al.\egroup
  }{2014}]{Liu2014}
Ziwei Liu, Ping Luo, Xiaogang Wang, and Xiaoou Tang.
\newblock Deep learning face attributes in the wild.
\newblock {\em arXiv:1411.7766}, 2014.

\bibitem[\protect\citeauthoryear{Mao \bgroup \em et al.\egroup
  }{2017}]{Mao2017}
Xudong Mao, Qing Li, Haoran Xie, Raymond~Y.K. Lau, Zhen Wang, and Stephen~Paul
  Smolley.
\newblock Least squares generative adversarial networks.
\newblock In {\em International Conference on Computer Vision (ICCV)}, 2017.

\bibitem[\protect\citeauthoryear{Mirza and Osindero}{2014}]{Mirza2014}
Mehdi Mirza and Simon Osindero.
\newblock Conditional generative adversarial nets.
\newblock {\em arXiv:1411.1784}, 2014.

\bibitem[\protect\citeauthoryear{Nowozin \bgroup \em et al.\egroup
  }{2016}]{Nowozin2016}
Sebastian Nowozin, Botond Cseke, and Ryota Tomioka.
\newblock f-gan: Training generative neural samplers using variational
  divergence minimization.
\newblock {\em arXiv:1606.00709}, 2016.

\bibitem[\protect\citeauthoryear{Perarnau \bgroup \em et al.\egroup
  }{2016}]{Perarnau2016}
Guim Perarnau, Joost van~de Weijer, Bogdan Raducanu, and Jose~M. {\'A}lvarez.
\newblock Invertible conditional gans for image editing.
\newblock {\em arXiv:1611.06355}, 2016.

\bibitem[\protect\citeauthoryear{Radford \bgroup \em et al.\egroup
  }{2015}]{Radford2015}
Alec Radford, Luke Metz, and Soumith Chintala.
\newblock Unsupervised representation learning with deep convolutional
  generative adversarial networks.
\newblock {\em arXiv:1511.06434}, 2015.

\bibitem[\protect\citeauthoryear{Roth \bgroup \em et al.\egroup
  }{2017}]{Roth2017}
Kevin Roth, Aurelien Lucchi, Sebastian Nowozin, and Thomas Hofmann.
\newblock Stabilizing training of generative adversarial networks through
  regularization.
\newblock {\em arXiv:1705.09367}, 2017.

\bibitem[\protect\citeauthoryear{Silberman \bgroup \em et al.\egroup
  }{2012}]{Silberman2012}
Nathan Silberman, Derek Hoiem, Pushmeet Kohli, and Rob Fergus.
\newblock Indoor segmentation and support inference from rgbd images.
\newblock In {\em European Conference on Computer Vision (ECCV)}, 2012.

\bibitem[\protect\citeauthoryear{Tieleman}{2008}]{Tieleman2008}
Tijmen Tieleman.
\newblock Training restricted boltzmann machines using approximations to the
  likelihood gradient.
\newblock In {\em International Conference on Machine Learning (ICML)}, 2008.

\bibitem[\protect\citeauthoryear{Tzeng \bgroup \em et al.\egroup
  }{2017}]{Tzeng2017}
Eric Tzeng, Judy Hoffman, Kate Saenko, and Trevor Darrell.
\newblock Adversarial discriminative domain adaptation.
\newblock In {\em Computer Vision and Pattern Recognition (CVPR)}, 2017.

\bibitem[\protect\citeauthoryear{Wang and Gupta}{2016}]{Wang2016}
Xiaolong Wang and Abhinav Gupta.
\newblock Generative image modeling using style and structure adversarial
  networks.
\newblock In {\em European Conference on Computer Vision (ECCV)}, 2016.

\bibitem[\protect\citeauthoryear{Wang \bgroup \em et al.\egroup
  }{2017}]{Wang2017}
Chaoyue Wang, Chaohui Wang, Chang Xu, and Dacheng Tao.
\newblock Tag disentangled generative adversarial networks for object image
  re-rendering.
\newblock In {\em International Joint Conference on Artificial Intelligence
  (IJCAI)}, 2017.

\bibitem[\protect\citeauthoryear{Yu and Grauman}{2014}]{Yu2014}
Aron Yu and Kristen Grauman.
\newblock {F}ine-{G}rained {V}isual {C}omparisons with {L}ocal {L}earning.
\newblock In {\em Computer Vision and Pattern Recognition (CVPR)}, 2014.

\bibitem[\protect\citeauthoryear{Zhu \bgroup \em et al.\egroup
  }{2016}]{Zhu2016}
Jun-Yan Zhu, Philipp Kr{\"a}henb{\"u}hl, Eli Shechtman, and Alexei~A. Efros.
\newblock Generative visual manipulation on the natural image manifold.
\newblock In {\em European Conference on Computer Vision (ECCV)}, 2016.

\bibitem[\protect\citeauthoryear{Zhu \bgroup \em et al.\egroup
  }{2017}]{Zhu2017}
Jun-Yan Zhu, Taesung Park, Phillip Isola, and Alexei~A. Efros.
\newblock Unpaired image-to-image translation using cycle-consistent
  adversarial networks.
\newblock In {\em International Conference on Computer Vision (ICCV)}, 2017.

\end{thebibliography}

\end{document}